\pdfoutput=1

\documentclass[11pt]{article}

\usepackage[final]{acl}

\usepackage{times}
\usepackage{latexsym}
\usepackage[autostyle, english = american]{csquotes}
\MakeOuterQuote{"}

\usepackage[T1]{fontenc}

\usepackage[utf8]{inputenc}

\usepackage{microtype}

\usepackage{inconsolata}

\usepackage{graphicx}
\usepackage{amsmath}
\title{Tuning Into Bias: A Computational Study of Gender Bias in Song Lyrics}

\author{
 \textbf{Danqing Chen$^*$},
 \textbf{Adithi Satish$^*$},
 \textbf{Rasul Khanbayov$^*$},
 \textbf{Carolin M. Schuster},
 \textbf{Georg Groh}
 \\
\\
 Technical University of Munich \\ Munich, Germany,
\\
 \small{
    {\{chen.danqing, adithi.satish, rasul.khanbayov, carolin.schuster\}@tum.de, grohg@in.tum.de
    }
    }
 }

\begin{document}
\maketitle
\def\thefootnote{*}\footnotetext{These authors contributed equally to this work}\def\thefootnote{\arabic{footnote}}

\begin{abstract}
The application of text mining methods is becoming increasingly prevalent, particularly within Humanities and Computational Social Sciences, as well as in a broader range of disciplines. This paper presents an analysis of gender bias in English song lyrics using topic modeling and bias measurement techniques. Leveraging BERTopic, we cluster a dataset of 537,553 English songs into distinct topics and analyze their temporal evolution. Our results reveal a significant thematic shift in song lyrics over time, transitioning from romantic themes to a heightened focus on the sexualization of women. Additionally, we observe a substantial prevalence of profanity and misogynistic content across various topics, with a particularly high concentration in the largest thematic cluster. To further analyse gender bias across topics and genres in a quantitative way, we employ the Single Category Word Embedding Association Test (SC-WEAT) to calculate bias scores for word embeddings trained on the most prominent topics as well as individual genres. The results indicate a consistent male bias in words associated with intelligence and strength, while appearance and weakness words show a female bias. Further analysis highlights variations in these biases across topics, illustrating the interplay between thematic content and gender stereotypes in song lyrics.
\end{abstract}

\section{Introduction}

\textbf{\textit{Disclaimer: Lyrics in the dataset may include explicit or vulgar language, which is inherently reflected in the topic labels generated by the BERTopic model. This does not represent the views or opinions of the authors.}} \\



Music is integrally tied with gender identity, where lyrics, melodies, and performance styles can reflect and shape societal perceptions of gender roles, stereotypes, and experiences~\cite{flynn2016objectification, colley2008young, alexander1999gender}. Through lyrics, artists have a way of expressing their emotions and discussing unique themes. While these themes often span a wide variety of issues, they can also propagate dangerous stereotypes and objectification~\cite{rasmussen2017girl, Sexualization2011Hall, Undressing2019Cynthia, From2017Smiler}, pointing out the need to critically examine these gender biases that can occur in lyrics.

\begin{figure}
    \centering
    \hspace{-4mm}
    \includegraphics[height=6.6cm]{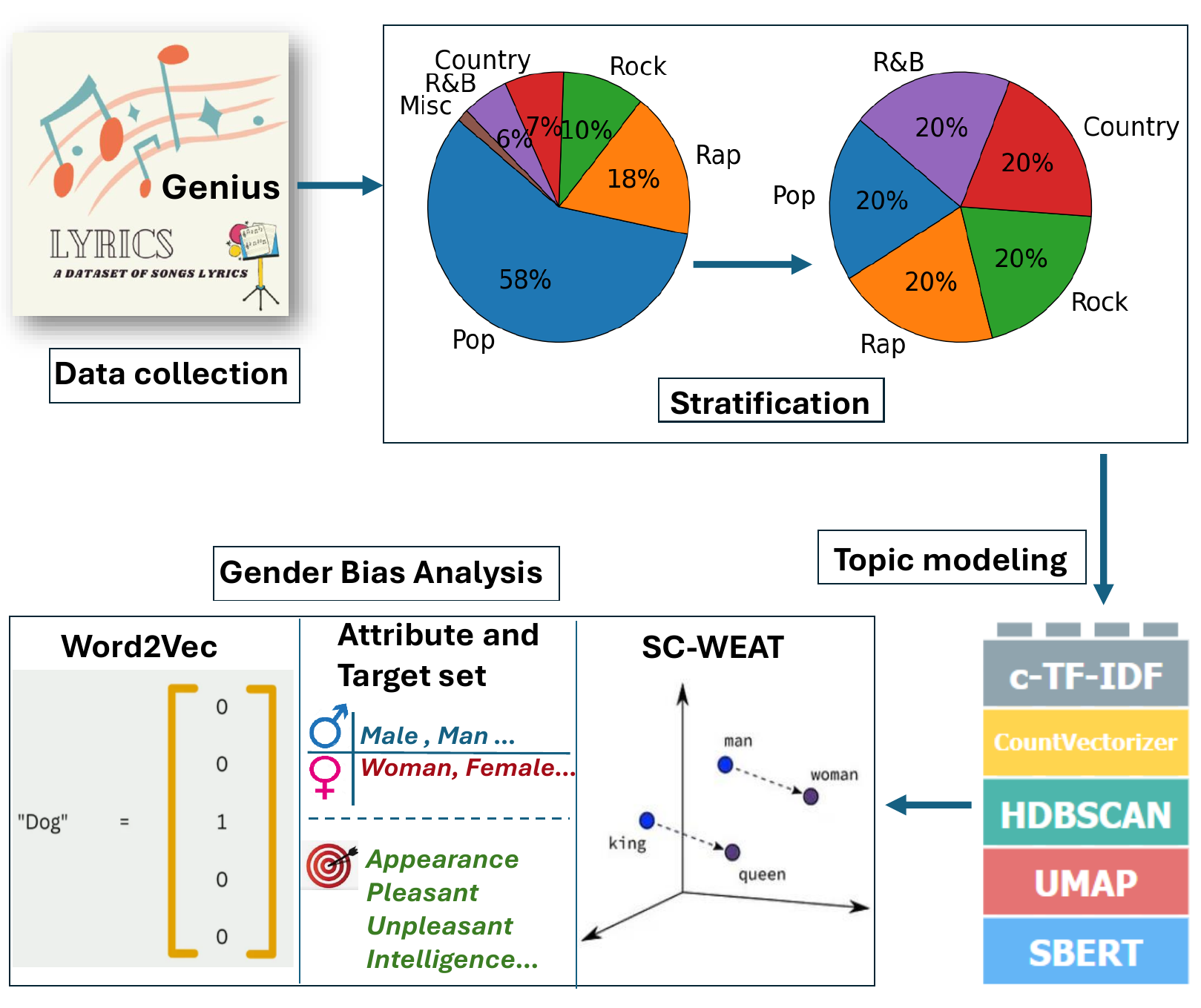}
    \caption{Detailed workflow including data collection, topic modeling, and SC-WEAT.}
    \label{fig:paper_pipeline}
\end{figure}


Natural Language Processing (NLP) techniques provide a robust framework for analyzing song lyrics by leveraging their underlying textual structure to extract thematic patterns and gender-associated linguistic representations~\cite{betti2023large}. In particular, word embeddings~\cite{bengio2000neural}, which encode lexical items as dense, high-dimensional vectors within a continuous space, have been shown to effectively capture and encode latent linguistic biases that align with human cognitive associations~\cite{Caliskan2017semantics, qin2023stereotype}. This representational property renders word embeddings a powerful computational tool for systematically quantifying and analyzing gender biases embedded within lyrical discourse~\cite{quantifyinggendeRBias}.

While previous research has primarily analyzed gender bias at the artist level by comparing the lyrics of songs performed by male and female artists~\cite{anglada2021popular, betti2023large, boghrati2023quantifying}, this study does not differentiate based on the artist's gender. Instead, we focus solely on examining bias within the lyrics themselves. By integrating topic modelling with quantitative bias measurement, this approach facilitates a granular analysis of gender bias across themes and genres, utilizing NLP to bridge the gap in Humanities and Social Sciences to analyze complex text-based artefacts and their sociocultural implications.

Topic modeling is a powerful technique for uncovering the underlying themes within a corpus, such as song lyrics in our study~\cite{kleedorfer2008oh}. In this paper, we employ BERTopic~\cite{grootendorst2022bertopic}, a state-of-the-art topic modeling method, to analyze persistent lyrical themes across various genres and examine their evolution over multiple decades. This approach enables us to uncover critical insights, including the increasing sexualization of women in song lyrics over time and the notable prevalence of profanity, particularly in rap music. While the topic model provides a broad overview of the gender bias in lyrics, we also take a more fine-grained look into this bias by applying the SC-WEAT analysis to quantify it and evaluate the associations of specific target word sets with gender-related attributes~\cite{mikolov2013efficient, Caliskan2017semantics}. Our major contributions, as depicted in the workflow diagram in Figure~\ref{fig:paper_pipeline}, are:
\begin{itemize}
    \item Conducting topic analysis on a stratified sample of song lyrics to identify cross-genre themes, recurrent topics, and the historical evolution of gender bias.
    \item Evaluating the prevalence and variation of gender bias in lyrics quantitatively across topics and genres through the computation of SC-WEAT scores.
\end{itemize}

\section{Related Work}



The intersection of music and natural language processing (NLP) has been the focus of extensive research, encompassing tasks such as mood classification, music transcription, lyrics and melody generation, among others~\citep{laurier2008multimodal, benetos2018automatic, chen2020melody, yu2021conditional}. Music — and, by extension, lyrics — constitutes a valuable resource for investigating underlying societal dynamics, particularly in the context of gender stereotypes and objectification~\citep{flynn2016objectification, bretthauer2007feminist, smiler2017want, quantifyinggendeRBias}.

Previous research has demonstrated that word embeddings are inherently susceptible to capturing and, in some cases, amplifying the social biases present in the data from which they are derived~\cite{hovy2021five}. A well-known example provided by~\citet{bolukbasi2016man} illustrates that the word embedding for ``man'' is more closely associated with ``programmer,'' while ``woman'' is linked to ``homemaker.'' Similarly, the findings of~\citet{durrheim2023using} and~\citet{zhao2019gender} reveal that word embeddings encode implicit cultural and gender biases, even when such biases are not explicitly stated in the source data. This body of work highlights the critical importance of examining and addressing biases embedded in linguistic representations, especially when applied to cultural artifacts such as song lyrics.

In our paper, we quantify this gender bias using an extension of the Word Embedding Association Test (WEAT), the Single Category WEAT score (SC-WEAT)~\citep{Caliskan2017semantics, charlesworth2021gender, betti2023large}. The SC-WEAT score is also used by~\citet{betti2023large} and ~\citet{boghrati2023quantifying} to analyze the nature of gender bias in lyrics and the differences across artist genders. However, we expand on this approach by using topic modeling to identify popular and intriguing topics. We then analyze the gender bias in the lyrics on a per-topic as well as per-genre basis, aiming to uncover how this bias may vary across different themes.

Topic modeling is a widely used technique for clustering documents to summarize or classify them, enabling the identification of underlying social patterns within the data~\cite{egger2022topic}. When applied to song lyrics, it serves as an effective approach for uncovering recurring themes~\citep{kleedorfer2008oh, fell2023wasabi, devi2020exploiting, karamouzi2024historical}. While Latent Dirichlet Allocation (LDA) remains one of the most common methods for topic modeling, recent findings by~\citet{gan2023experimental} demonstrate that BERTopic, introduced by~\citet{grootendorst2022bertopic}, outperforms traditional approaches by producing more distinctive and interpretable clusters. BERTopic has also been successfully applied in gender and social science research. For example,~\citet{nakajima-wickham-2023-girlbosses} utilized the algorithm to examine gender expectations on social media and their influence on suicidal ideation. This demonstrates BERTopic's utility in the clustering of categories that are meaningful to societal and cultural dynamics.

\section{Experimental Setup}
\subsection{Data}

The dataset used for the lyric analysis is a combination of song metadata from the WASABI Song Corpus created by~\citet{fell2023wasabi}, and English lyrical content from Genius Song Lyrics~\footnote{\url{https://www.kaggle.com/datasets/carlosgdcj/genius-song-lyrics-with-language-information}}. Our lyrics dataset includes data as recent as 2022 extracted from Genius, an online platform where users can upload and explain songs, poems, and even books but primarily focus on songs.

The final dataset consists of 537,553 song lyrics across five main genres and an additional miscellaneous category as described in Table~\ref{tab:datasetsummary}. 


\renewcommand{\arraystretch}{1.1} 
\begin{table}[h]
\centering
\begin{tabular}{l c c c}
\hline
\textbf{Genre} & \textbf{Counts (\% of dataset)}\\
\hline
Pop & 311,085 (58\%) \\
Rap & 94,234 (18\%) \\
Rock & 54,560 (10\%) \\
Country & 39,078 (7\%) \\
R\&B & 30,747 (6\%) \\
Misc & 7,849 (1\%) \\
\hline
\end{tabular}
\caption{Counts of songs across genres in the dataset.}
\label{tab:datasetsummary}
\end{table}



\subsection{Topic Modeling with BERTopic}
\label{bertopic}
BERTopic leverages transformers to create clusters, providing more interpretable topic representations compared to traditional methods~\cite{grootendorst2022bertopic}. The algorithm creates topics in four steps, which involve (i) transforming the documents into embeddings using a pre-trained language model, (ii) reducing their dimensionality, (iii) clustering and finally, (iv) deriving the topic representations from these clusters using a class-based version of TF-IDF. For our analysis, we use the default configuration of BERTopic, which utilizes (i) all-Mini-LM-L6-V2~\footnote{\url{https://huggingface.co/sentence-transformers/all-MiniLM-L6-v2}}, (ii) UMAP, (iii) HDBSCAN and (iv) c-TF-IDF for the four steps mentioned above~\footnote{\url{https://maartengr.github.io/BERTopic/algorithm/algorithm.html}}.

BERTopic leverages c-TF-IDF (class-based Term Frequency-Inverse Document Frequency) to represent topics by weighting words based on their importance within a topic rather than across the entire corpus~\cite{grootendorst2022bertopic}. This approach emphasizes words that are not only frequent within a given topic but also capable of distinguishing that topic from others in the dataset. 
To optimize computational resources while preserving dataset representativeness, we train the BERTopic model on a stratified sample comprising 20,000 songs per genre and 7,849 ``misc'' entries. The model then predicts topic labels for the full corpus, which are subsequently analyzed for gender bias using SC-WEAT scores.

\subsection{Bias Measurements - SC-WEAT}\label{scweat_details}
To analyze gender bias in lyrics, we quantify the bias by training word embeddings from scratch to compute their association scores, using an extension of the original WEAT score~\cite{Caliskan2017semantics, charlesworth2021gender}, called the SC-WEAT score, which quantifies the relationship between a set of target words and two sets of attribute words~\cite{betti2023large}.

SC-WEAT Score Formula: The association strength is calculated using the formula below, as proposed by~\citet{Caliskan2017semantics} and used by~\citet{betti2023large}:

\begin{equation}
\begin{aligned}
    s(w, A, B) &= \text{mean}_{a \in A} \cos (\vec{w}, \vec{a}) \\ &- \text{mean}_{b \in B} \cos(\vec{w}, \vec{b})
\end{aligned}
\end{equation}

\begin{equation}
\begin{aligned}
    \text{SCWEAT}(X, A, B) &= \sum_{x \in X} s(x, A, B)
\end{aligned}
\end{equation}

\begin{equation}
\begin{aligned}
    d &= \frac{\text{mean}_{x in X}s(x, A, B)}{\text{stddev}_{x in X}s(x, A, B)}
\end{aligned}
\end{equation}


The cosine similarity \textit{s(w,A,B)} is the difference between the mean cosine similarity of the word vector \textit{w} to vectors in attribute sets \textit{A} and \textit{B}, respectively. The differential association, or effect size, is the normalized SC-WEAT score.

To compute SC-WEAT scores, we train Word2Vec embeddings for each genre and the top topic within each genre. Static embeddings, such as Word2Vec, are well-suited for analyzing aggregate biases within the data~\cite{Caliskan2017semantics, betti2023large}. As the objective is to examine gender bias inherent in the dataset rather than the model itself, Word2Vec—trained from scratch—is more appropriate than contextual models like BERT~\cite{mikolov2013efficient}.

We define six target sets, curated by~\citet{Caliskan2017semantics} and~\citet{chaloner2019measuring}, which are used by~\citet{betti2023large}, in addition to two attribute sets for male and female characteristics, respectively (see Table~\ref{tab:target_set}). The SC-WEAT scores are calculated for each of these target sets using the aforementioned formula for each embedding model. A negative SC-WEAT score indicates a higher similarity towards the female attribute set, whereas a positive score indicates a higher similarity towards the male attribute set. The magnitude of the effect size indicates the strength of the respective bias.

\begin{table}[]
    \centering
    \scalebox{0.9}{
    \renewcommand{\arraystretch}{1.2} 
    \begin{tabular}{p{0.25\linewidth} | p{0.75\linewidth}}
    \hline
    \textbf{Target Set} & \textbf{Examples of words in the word sets} \\
    \hline
    Pleasant & ``joy'', ``wonderful'', ``love'', ``peace'' \\
    Unpleasant & ``terrible'', ``hatred'', ``nasty'', ``kill'' \\
    Appearance &``thin'',``gorgeous'',``fat'',``pretty'' \\
    Intelligence &``intelligent'',``genius'',``brilliant'' \\
    Strength &``bold'',``leader'',``strong'',``power'' \\
    Weakness &``loser'',``failure'',``weak'',``follow'' \\
    \hline
    \hline
    \textbf{Attribute Set} & \textbf{Examples of words in the word sets} \\
    \hline
    Female &``girl'',``her'',``woman'',``girlfriend'' \\
    Male &``boy'',``him'',``man'',``boyfriend'' \\
    \hline
    \end{tabular}}
    \caption{Examples of target and attribute sets used for SC-WEAT analysis. The full lists of words, curated by~\citet{betti2023large}, can be found in \autoref{tab:target_set_appendix} and \autoref{tab:attribute_set_appendix} in the Appendix} 
    \label{tab:target_set}
\end{table}







\section{Results \& Discussion}

\subsection{Topic Analysis}
\label{topic_analysis}

The BERTopic model identifies a total of 541 topics, with 1.5\% of documents classified as outliers. Figure \ref{fig:top5crossgenretopics} illustrates the most salient topics along with their genre distributions, representing the genre composition of songs assigned to each topic label, where each label is generated based on the most representative terms, constructed using the top three words with the highest c-TF-IDF values. 

While the figure shows the composition of the top topics in each genre, it reveals the dominant influence of pop in other genres as well.  For instance, in addition to the top topic within pop, the top topics in country (\textit{``tears\_heart\_wish''}), R\&B (\textit{``body\_girl\_baby''}) and rock (\textit{``ayy ayy\_change\_long\_sentiment''}) are also largely shaped or consist of pop songs. This indicates greater thematic diversity of pop songs, whereas rap exhibits a strong thematic concentration, with 89.2\% of songs in \textit{"nigga\_niggas\_bitch"} belonging to the rap genre. While pop is the most prevalent genre in the dataset (see Table~\ref{tab:datasetsummary}), this imbalance is mitigated by the stratified sampling approach outlined in Section~\ref{bertopic}, ensuring a more balanced genre representation in the analysis.

\begin{figure*}[h!]
    \centering
    \includegraphics[width=\textwidth]{figures/topic_2_show_9_last.pdf} %
   \caption{Distribution of the top topic in each genre, with (n) representing the number of songs associated with that topic. As shown, the top topic in each genre often includes a significant proportion of songs from other genres, indicating genre overlap in topic composition.} 
    \label{fig:top5crossgenretopics}
\end{figure*}

\begin{figure*}
    \centering
    \includegraphics[scale=0.376]{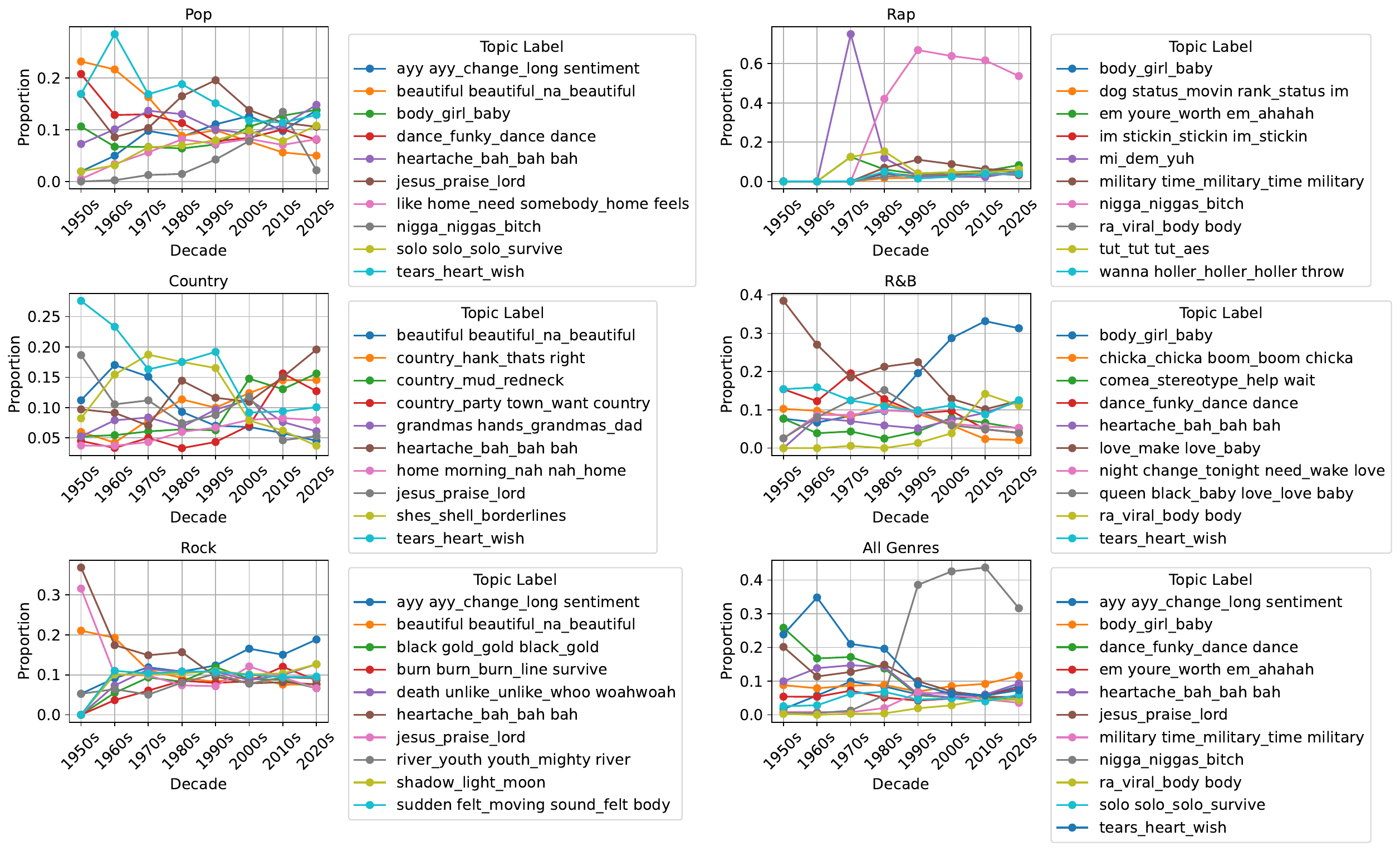}
    \caption{Development over time of top 10 topics in each genre and overall; decline from 2010 to 2020 can be explained by the yet still limited data for the 2020s.}
    \label{fig:topicdevelopmentoverdecades}
\end{figure*}

Despite the prevalence of pop music in the dataset, Figure \ref{fig:topicdevelopmentoverdecades} shows that the most prominent topic in rap, \textit{"nigga\_niggas\_bitch"}, has the highest frequency across all genres and emerged predominantly in the 1990s. Analyzing the distribution of top topics within each genre highlights a stark disparity: the top topic in pop accounts for only 1.77\% of all pop songs, whereas in rap, the top topic represents 37.88\% of the genre. This significant concentration indicates the dominant popularity and thematic specificity of this topic within rap, accounting for a substantial portion of the dataset.

\begin{figure}[ht!]
    \centering
        \centering
        \includegraphics[scale=0.3]{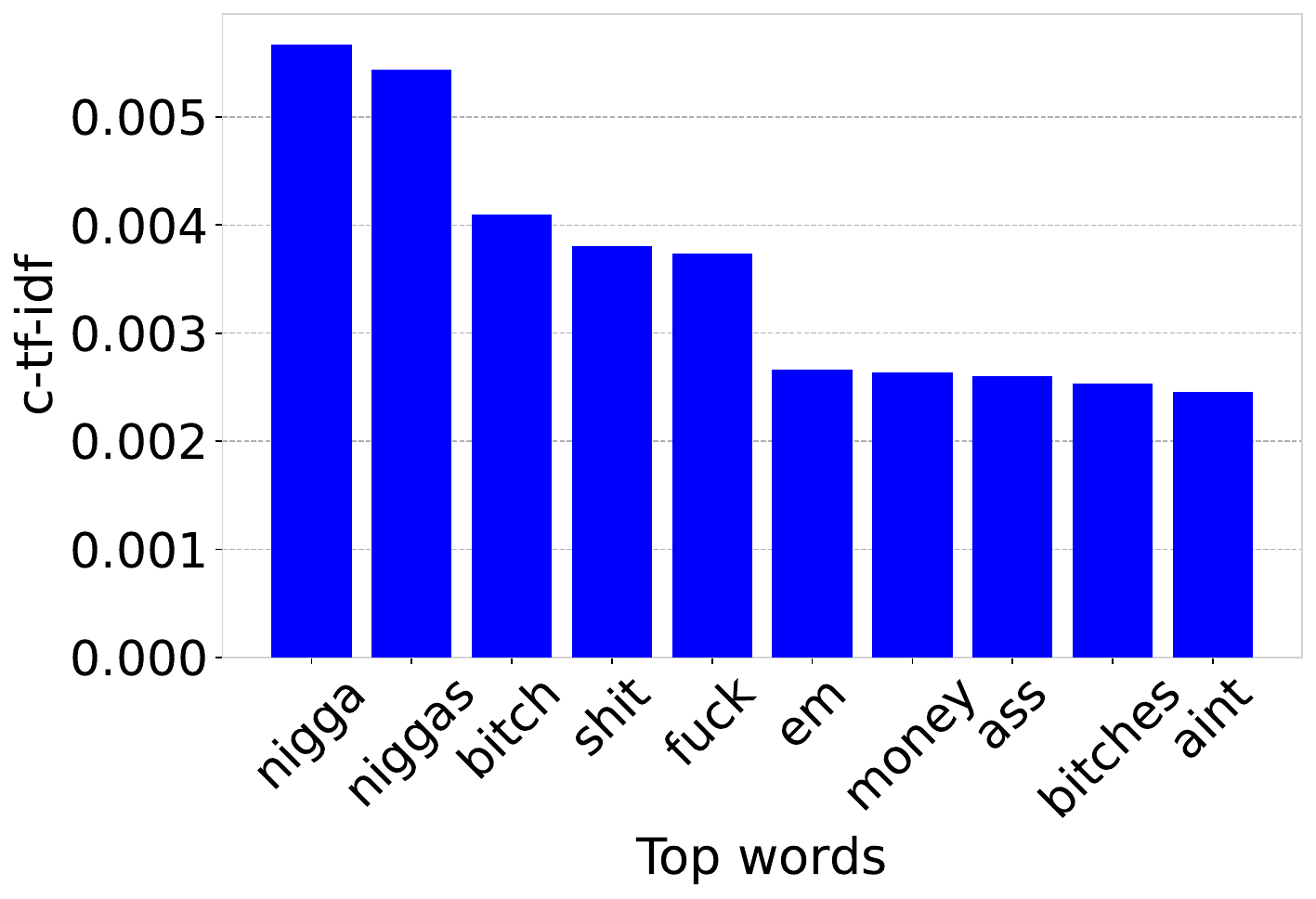}
        \caption{c-TF-IDF score for the overall top topic: \textit{"nigga\_niggas\_bitch"}}
        \label{fig:nigga_niggas_bitch} 
\end{figure}

This pronounced disparity emphasizes the distinctive narrative centrality of the top topic in rap compared to pop, necessitating a more detailed investigation into its linguistic and cultural characteristics. An analysis of the lyrics within this topic reveals a frequent occurrence of vulgar language and profanity, as evident from the c-TF-IDF scores (see Figure \ref{fig:nigga_niggas_bitch}). These observations highlight the thematic uniqueness of rap and underline the importance of further examining the social and cultural implications embedded within its lyrical content.

A detailed qualitative analysis of the lyrics within this topic, exemplified by tracks such as \textit{Big L's 7 Minute Freestyle} and \textit{Eminem's Kill You}, reveals a prevalent use of explicit and coarse language. Notable lyrical excerpts, including \textit{``F*ck love / All I got for hoes is hard d*ck and bubblegum' and 'Slut, you think I won't choke no whore / Til the vocal cords don't work in her throat no more?!''}, exemplify this linguistic trend. These findings align with the argument presented by~\citet{evadewi2018analysis}, who contend that rap music lyrics are distinguished by the frequent incorporation of vulgar and explicit language, setting them apart from other English-language musical genres. Furthermore, a quantitative analysis of word frequency within this topic and across rap lyrics underscores the recurrent presence of misogynistic terminology, which serves to reinforce negative gender stereotypes and perpetuate discriminatory narratives. In particular, derogatory terms such as 'bitches,' 'sluts,' and 'hoes' frequently appear in reference to women, reflecting broader patterns of gendered linguistic bias within this lyrical subdomain. This observation is further corroborated by \citet{doi:10.1177/0021934704274072} and \citet{Gronevik658843}, who highlight that such ideologies manifest through a spectrum of expressions, from subtle insinuations to obvious stereotypical representations and defamatory language within rap lyrics. Additionally, the higher prevalence of misogyny and profanity in rap lyrics, compared to other genres, aligns with the findings of \citet{Undressing2019Cynthia}, who document similar patterns in their comprehensive analyses.

Furthermore, \citet{From2017Smiler} also documented the evolution of music content over time, shifting from themes related to romantic relationships to an increase in references to sexual behaviour and objectified bodies, as evidenced in the topics in rap. This is also proven in our findings that in the top topics across successive decades, the following topics appear as trending:  \textit{"wonderful\_sweeter years\_sweeter"}, spanning from the 1950s to the 1960s, (due to fewer occurrences of this topic, it does not feature in Figure \ref{fig:topicdevelopmentoverdecades}), \textit{"tears\_heart\_wish"}, from 1960s to the 1980s, and \textit{"nigga\_niggas\_bitch"} from 1980s to 2020s. This observation is consistent with the results reported by  \citet{Sexualization2011Hall}, who found that when comparing lyrics from 2009 to those from 1959, the occurrence of sexualized content in 2009 was over three times higher.

\subsection{SC-WEAT Analysis}

\begin{figure*}[h!]
        \centering
        \vspace{-4mm}
        \includegraphics[scale=0.35]{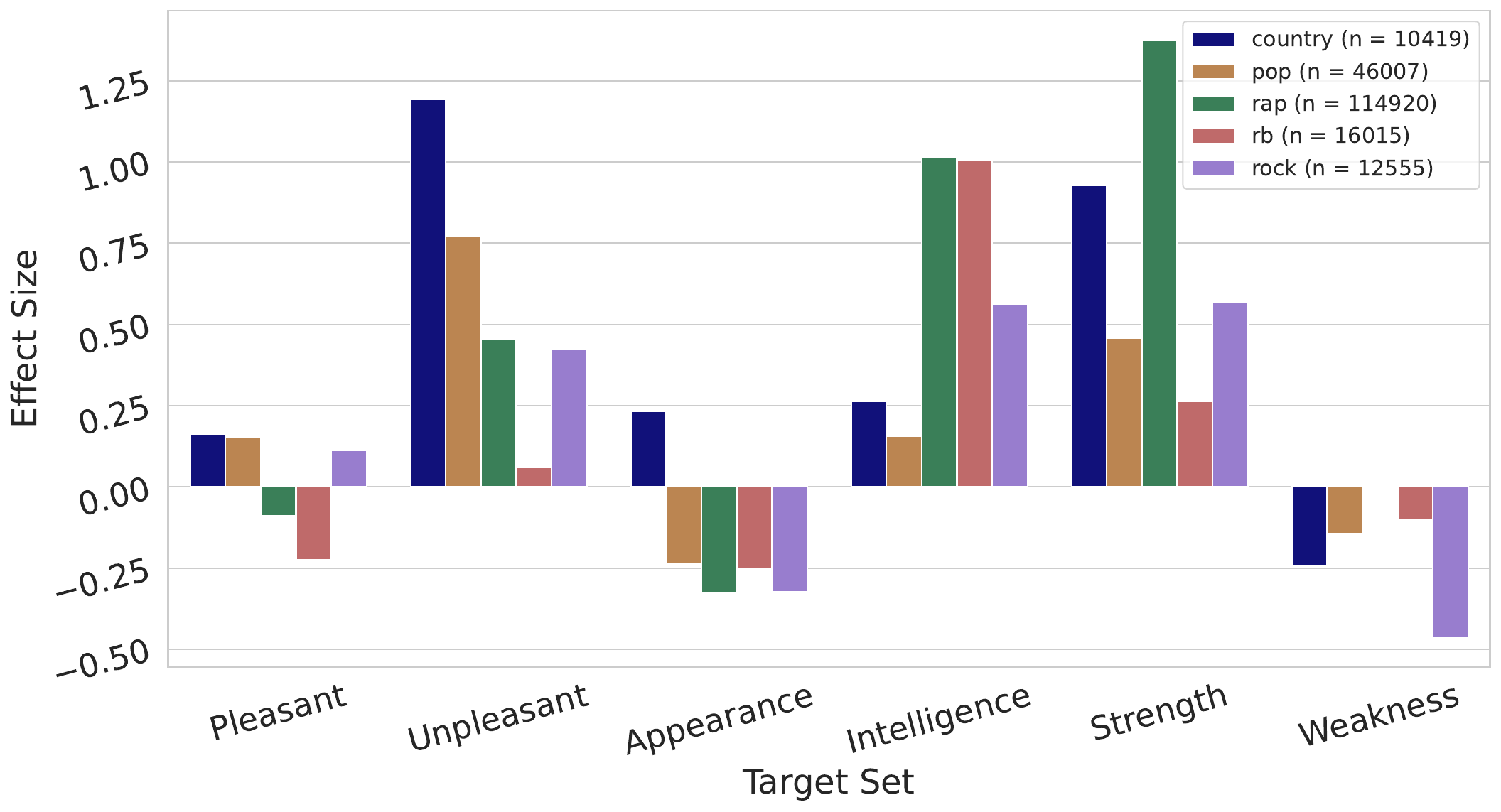}
        \caption{The SC-WEAT effect size of the target sets in each genre. A positive score indicates male bias, whereas a negative score indicates female bias, and n represents the number of word vectors for each genre.}
        \label{fig:scweatavgbias}
        \vspace{-5mm}
\end{figure*}

\begin{figure}[h]
        \centering
        \includegraphics[scale=0.3]{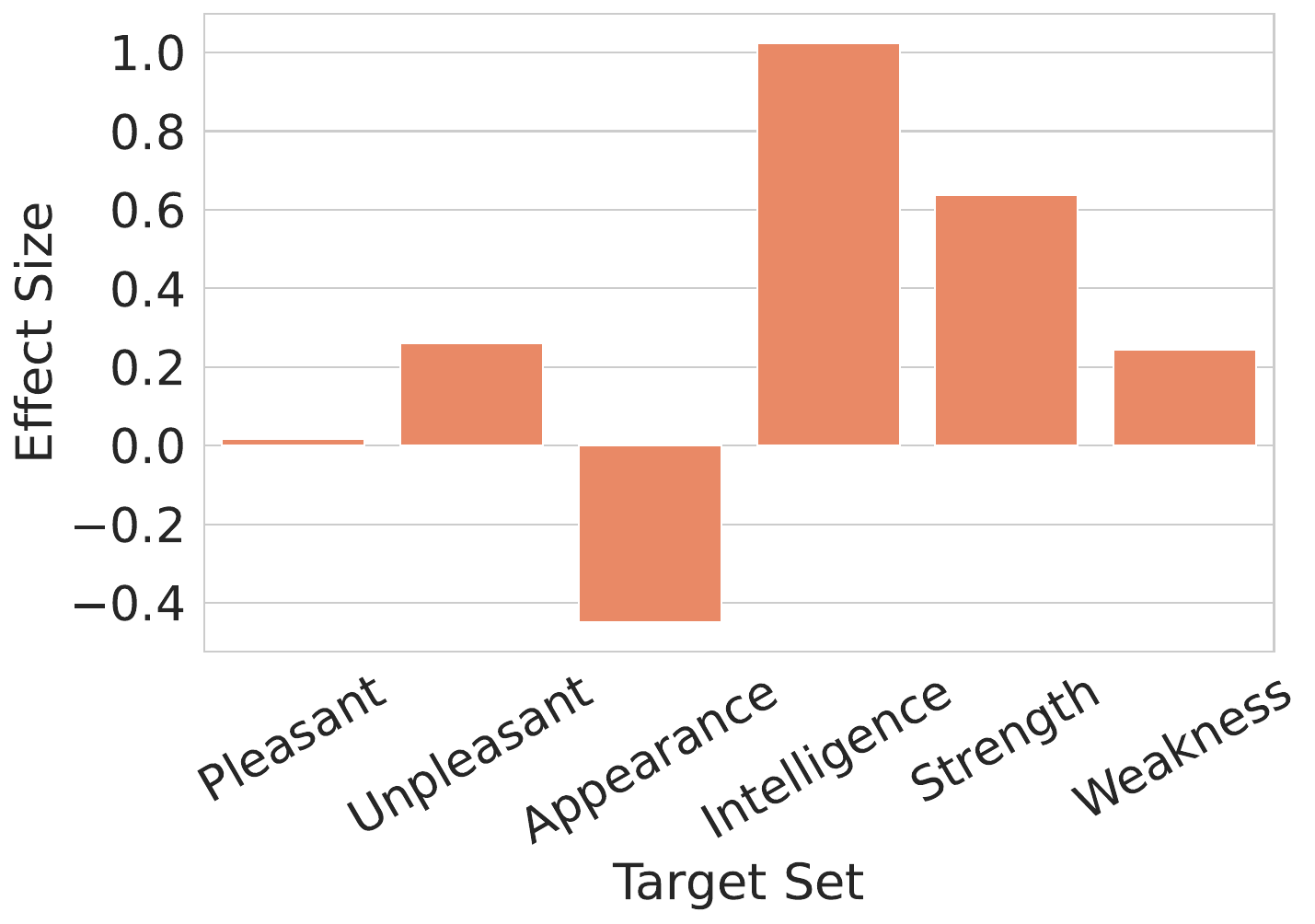}
        \caption{SC-WEAT score for the top topic: \textit{"nigga\_niggas\_bitch"}. A positive score indicates male bias, whereas a negative score indicates female bias, and n is the number of word vectors.}
        \label{fig:nnb_scweat} 
    \label{fig:combined_tfidf_scores} 
\end{figure}

\begin{figure*}[ht!]
    \centering
        \centering
        \includegraphics[clip, scale=0.36]{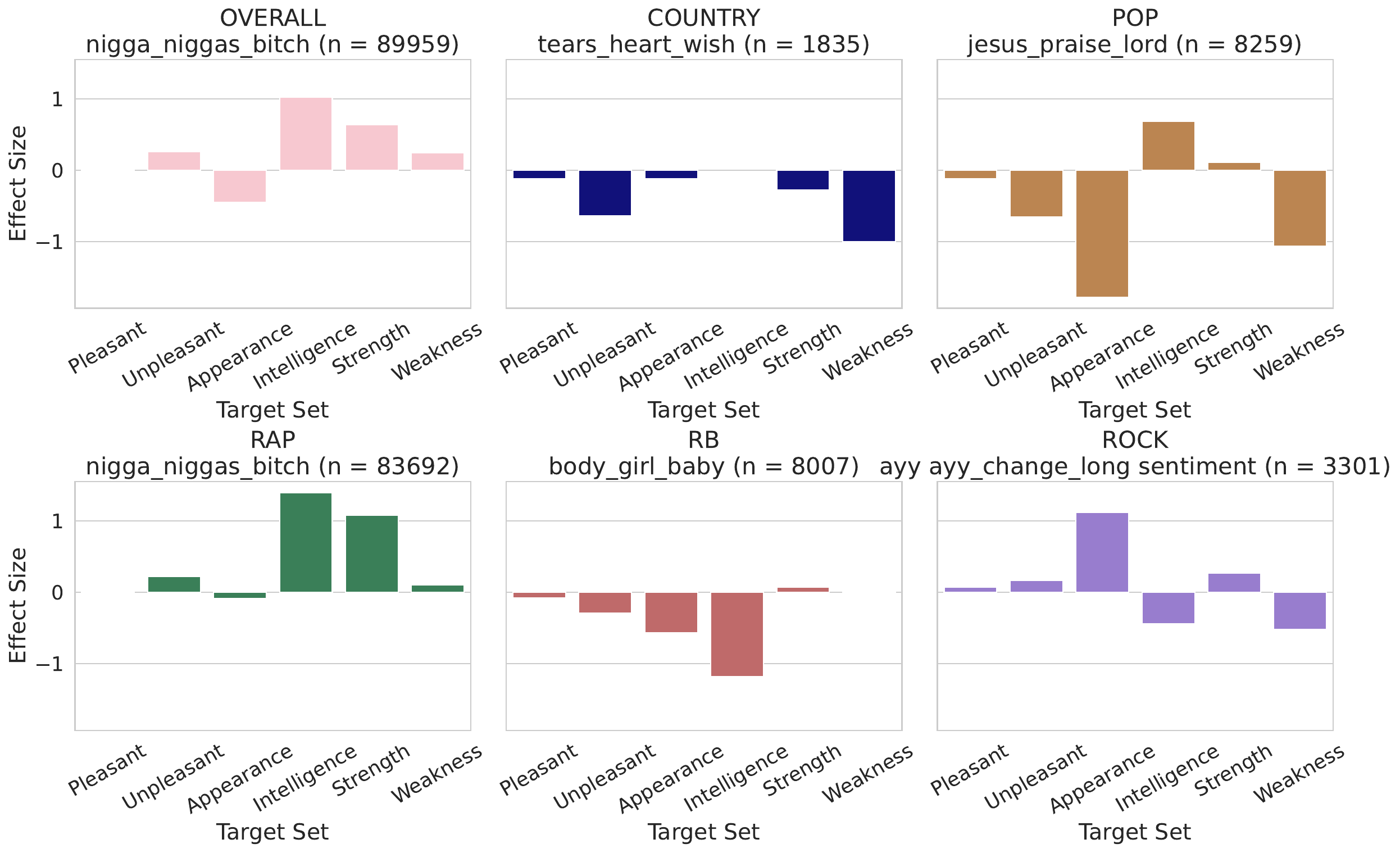}
        \caption{Comparison of SC-WEAT bias plots of the top topics in each genre. A positive score indicates male bias, whereas a negative score indicates female bias, and n is the number of word vectors.}
        \label{fig:scweat_toptopic}
\end{figure*}

Employing these topics as grouping indicators, we analyze gender bias in the lyrics by calculating the SC-WEAT scores, grouped by genre, as shown in Figure \ref{fig:scweatavgbias}. We observe no common trend in any genre to be male or female-biased overall; instead, they show variations in each target set.


We observe that Unpleasant, Intelligence, and Strength words exhibit positive SC-WEAT scores across all genres, with notably higher effect sizes in rap and country. This indicates that these target sets are more closely associated with male attributes on average, reflecting a pronounced male bias. These findings align with prior research by~\citet{betti2023large}, which highlights the strong association between Strength words and male nouns or names. Furthermore, the observed male bias aligns with prior research indicating that men are more frequently associated with attributes related to competence, such as 'smart,' 'strong,' and 'brave,' in contrast to women~\cite{quantifyinggendeRBias, boghrati2023quantifying}.

A systematic analysis of female bias within song lyrics reveals that the Weakness target set consistently exhibits negative SC-WEAT scores across multiple genres. This trend suggests that, in parallel with the stronger association of men with competence-related attributes, women are more frequently linked to concepts of weakness. Such linguistic patterns reinforce entrenched gender stereotypes, thereby perpetuating and amplifying gendered asymmetries in lyrical discourse.

This phenomenon aligns with prior findings by~\citet{liu-etal-2023-understanding}, which highlight the prevalence of gender stereotypes in media, such as the association of men with strength and women with appearance, particularly in contexts like video games. Similarly, the corpus-based study by~\citet{krasse2019corpus} on pop lyrics identifies a pronounced linguistic pattern wherein adjectives such as ``pretty,'' ``beautiful,'' ``ugly,'' and ``baby'' frequently precede female nouns. Our empirical analysis substantiates these findings, revealing that Appearance-related words consistently yield negative SC-WEAT scores across four out of five musical genres. This trend highlights the predominant linguistic association of women with attributes linked to physical appearance rather than intellectual or competence-related qualities. These results are consistent with prior research documenting the pervasive sexualization and objectification of women in song lyrics~\cite{flynn2016objectification, Sexualization2011Hall, karsay2019increasingly, rasmussen2017girl}, further illustrating how this cultural medium serves to reinforce and perpetuate traditional gender stereotypes.

For a more granular analysis, we compute SC-WEAT scores for the top topic in each genre and overall. Figure \ref{fig:nnb_scweat} visualizes the scores for the top overall topic (\textit{"nigga\_niggas\_bitch"}), where Appearance words exhibit a strong female bias, while Intelligence words show a marked male bias. These findings reinforce the gender divide and the objectification of women within this topic, as discussed in Section \ref{topic_analysis}.

Furthermore, Figure~\ref{fig:scweat_toptopic} illustrates that the biases associated with target sets vary across topics. Notably, Appearance words generally exhibit a female bias; however, in the topic \textit{"ayy ayy\_change\_long sentiment"}, they display a male bias, while Intelligence words show a female bias—contrasting with the overall trend observed in the rock genre (refer to Figure~\ref{fig:scweatavgbias}). These findings emphasize the importance of topic-specific analysis to capture the nuanced variations in biases across different topics, which might otherwise be obscured in genre-level aggregations.

\begin{figure*}
        \centering
        \includegraphics[clip, scale=0.37]{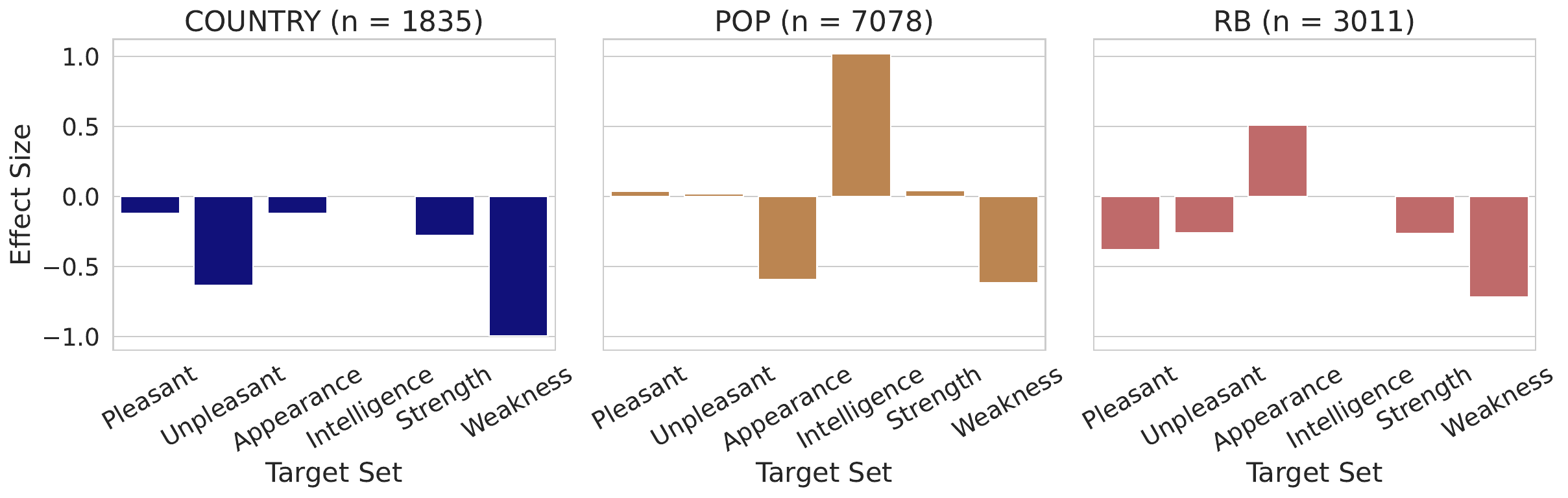}
        \caption{Comparison of SC-WEAT plots for \textit{"tears\_heart\_wish"} in the country, pop, and R\&B genres. A positive score indicates male bias, whereas a negative score indicates female bias, and n is the number of word vectors. }
        \label{fig:scweat_thw}
    
\end{figure*}

Moreover, certain prevalent topics that appear across multiple genres exhibit differing biases depending on the genre (see Figure \ref{fig:top5crossgenretopics}). For instance, the topic \textit{"tears\_heart\_wish"}, which is present in the country, pop, and R\&B genres, demonstrates distinct SC-WEAT scores for each genre, as shown in Figure \ref{fig:scweat_thw}. In the country genre, this topic consistently displays a female bias across all target sets, with Weakness words showing the strongest bias. These findings align with prior research by~\citet{rasmussen2017girl}, which observed that over half of the country songs analyzed reinforce stereotypical female gender roles and objectify women. This underscores the role of genre-specific contexts in shaping the gendered associations present in song lyrics.

Figure \ref{fig:scweat_thw} reveals that Weakness words consistently exhibit a female bias across the three genres analyzed, aligning with our broader observation that women are more frequently associated with weakness. Notably, Intelligence words in country and R\&B deviate from their average bias trends (see Figure \ref{fig:scweatavgbias}), as these genres typically display a strong male bias overall yet show negligible scores for this specific genre.

The influence of genre-specific dynamics is further highlighted by the behaviour of Appearance words in Figure \ref{fig:scweat_thw}. While Appearance words display a male bias in R\&B, they exhibit a female bias in pop, demonstrating how the same topic can exhibit divergent biases depending on the genre. These findings underscore the critical role of genre in shaping the gendered associations of recurring themes within song lyrics, emphasizing the need for a nuanced, genre-sensitive analysis to fully understand the interplay between thematic content and gender bias.

\section{Conclusion}

As a socio-cultural artefact, music offers insights into societal norms and biases, making it a valuable subject for computational analysis. This study leverages BERTopic, an advanced topic modeling technique, to identify thematic patterns and gender bias in song lyrics across five genres—country, pop, rap, R\&B, and rock—over 70 years. Using SC-WEAT, we quantify gender bias within these themes and explore how biases vary across topics and genres. By addressing the intersection of music, culture, and societal norms, our findings reveal the gendered narratives embedded in song lyrics and their evolution over time.

We employ a stratified sampling strategy for BERTopic model training to ensure balanced genre representation. The most dominant topic, \textit{"nigga\_niggas\_bitch"}, exhibits a high prevalence of misogynistic language and profanity, becoming particularly prominent in the 1990s despite the dataset spanning from the 1950s to the 2020s. In contrast, earlier dominant themes, such as \textit{"tears\_heart\_wish"} and \textit{"wonderful\_sweeter\_years\_sweeter"}, primarily reflect romantic and sentimental content. Over time, these themes shift toward heightened sexualization and explicit language, reflecting broader sociocultural and linguistic transformations in popular music, aligning with prior research on the increasing prevalence of sexualized and gendered language in song lyrics~\cite{Sexualization2011Hall, smiler2017want}.

The SC-WEAT analysis further examines the trends of sexualization and profanity previously identified through topic modeling. The results reveal implicit gender bias in song lyrics, with Weakness and Appearance words showing a female bias, while Intelligence and Strength words exhibit a male bias. The female bias in Appearance words supports observations on the sexualization of women in music~\cite{flynn2016objectification, Sexualization2011Hall, rasmussen2017girl}. The per-topic and per-genre analysis uncovers notable variations, with biases differing across themes and genres.

For instance, in the topic \textit{“tears\_heart\_wish,”} bias scores vary across genres: country exhibits a female bias across all target sets, while Intelligence words in pop and Appearance words in R\&B show a male bias. These results highlight the intersection of thematic content, genre, and gender bias, emphasizing the value of computational methods in analyzing sociocultural dynamics in song lyrics.

In conclusion, this study demonstrates the utility of integrating topic modeling with bias measurement techniques to analyze thematic structures in song lyrics and examine how these themes perpetuate implicit gender biases. By applying NLP methods to a significant sociocultural dataset, this work aligns with the growing demand in Digital Humanities and Social Sciences for tools that facilitate the analysis and interpretation of complex, non-standard textual data. Our approach highlights the potential of computational methods to address sociocultural questions, offering insights into how gender stereotypes are embedded in and perpetuated through lyrical content.

\section*{Limitations}

\textit{Language Limitations:} This study focuses exclusively on English-language songs, despite the multilingual content available on the Genius platform. Future research could expand to include songs in other languages, enhancing the scope and applicability of the findings.

\textit{Gender Classification:} This analysis treats gender as binary, overlooking the spectrum of gender identities. Future research should explore the full spectrum of gender diversity in music for more inclusive insights.

\textit{BERTopic Modeling:} A limitation of BERTopic, when applied to song lyrics analysis, is that it assigns a single topic per song, which does not account for songs that comprise different verses which may have different topics. 

\textit{Race and Gender:} In this paper, we look at the gender bias in lyrics independent of the race or gender of the artists, potentially neglecting their influence on the bias in the songs, especially in genres like rap. Future work could focus on integrating these aspects for a more detailed analysis of bias in music.

Addressing these limitations could significantly advance the field, offering an even more nuanced and comprehensive perspective on the intersection of music, culture, and societal norms.

\bibliography{acl_latex}

\appendix

\section{Appendix}
\label{sec:appendix}

\subsection{Data Cleaning}
\label{app_cleaning}
We gather the song metadata from the WASABI corpus~\footnote{\url{https://github.com/micbuffa/WasabiDataset}} and their respective lyrics information from the Genius Music Platform. Songs obtained from the Genius platform require preprocessing due to their unique format. Metadata associated with songs is typically enclosed within square brackets and embedded directly within the lyrical content. Additionally, the structure of the lyrics is generally preserved, resulting in entries that contain numerous newline characters. These characteristics may introduce challenges when parsing the data or preparing it for input into computational models, necessitating careful preprocessing to ensure consistency and usability. An example of the lyrics stored in the Genius dataset for ``Love Story'' by Taylor Swift: 

\begin{quote}
[Verse 1] \\
We were both young when I first saw you \\
I close my eyes and the flashback starts... 

[Pre-Chorus] \\
That you were Romeo, you were throwing pebbles \\
...
\end{quote}

\subsection{Analysis of genre popularity across decades}

\begin{figure}
    \centering
    \includegraphics[scale=0.20]{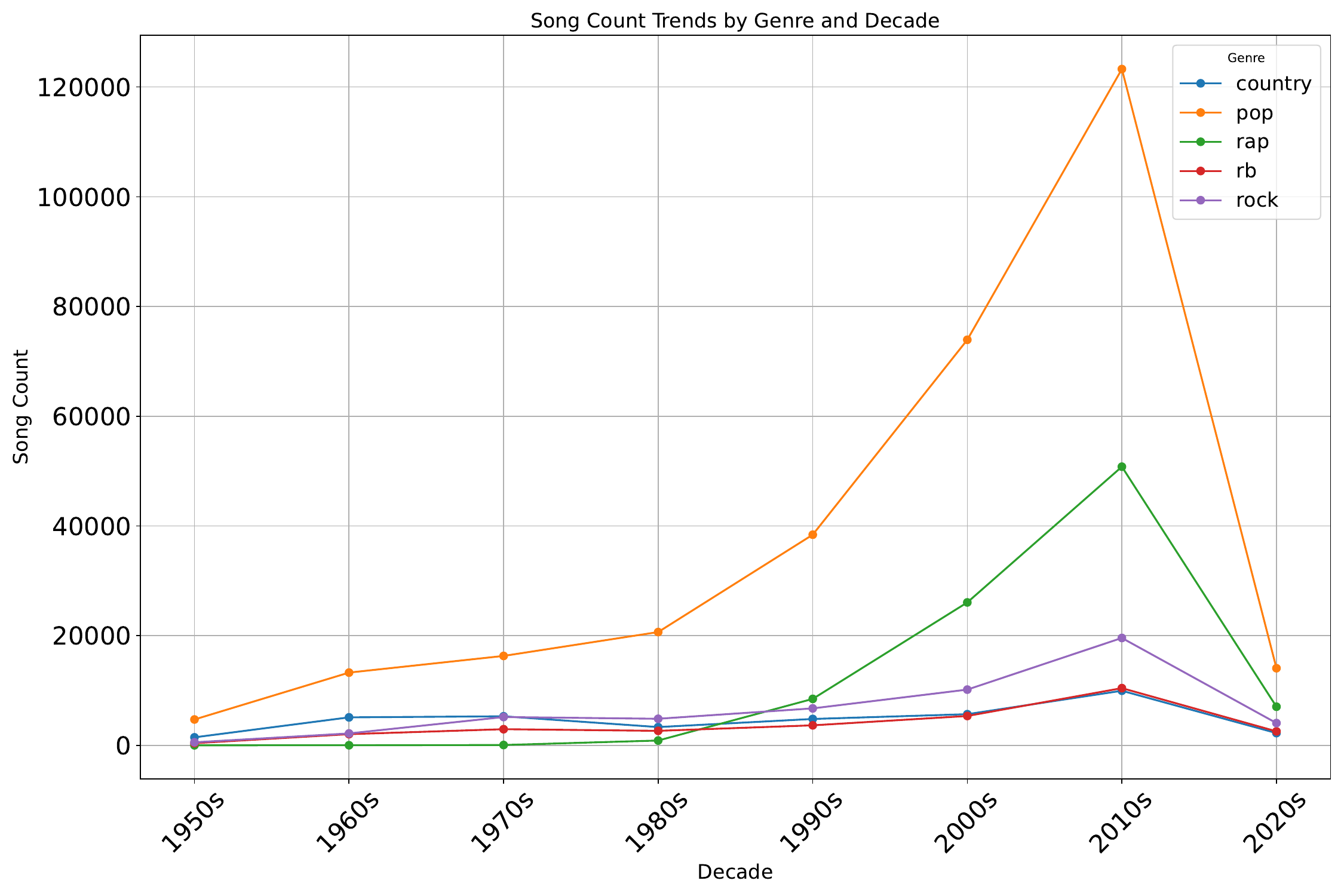}
    \caption{Genre trends over decades}
    \label{fig:genretrendsoverdecades}
\end{figure}

Figure \ref{fig:genretrendsoverdecades} presents a line chart illustrating the temporal evolution of genre popularity from the 1950s onward. In the early decades, country music demonstrates a higher relative prevalence compared to rap. However, a pronounced shift emerges in the 1990s, marked by a significant and rapid increase in the prominence of rap music. 




\subsection{Initial BERTopic Model}
The initial BERTopic model was trained on a randomly sampled subset of approximately 40,000 rows from the dataset. However, this approach resulted in an excessively high outlier rate, with over 50\% of entries (approximately 27,000 rows) classified as outliers. This necessitated computationally intensive post-processing steps for outlier reduction, ultimately rendering the model suboptimal for integration into the final analytical pipeline. To address this limitation, we employed a stratified sampling strategy, selecting 107,000 rows balanced across musical genres for model training, followed by transformation on the entire dataset. This revised approach led to a substantial improvement in model stability and representational fidelity, reducing the proportion of outliers to just 1.5\%. Consequently, this methodological refinement enhanced both the computational efficiency and the overall robustness of the topic modeling pipeline.


\subsection{Topic Label Analysis Using c-TF-IDF score from Bertopic model}
\label{topic_label_analysis_using_c_tf_idf_score_from_bertopic_model}

\begin{figure*}
    \centering    \includegraphics[width=\textwidth,height=2.0\textheight,keepaspectratio]{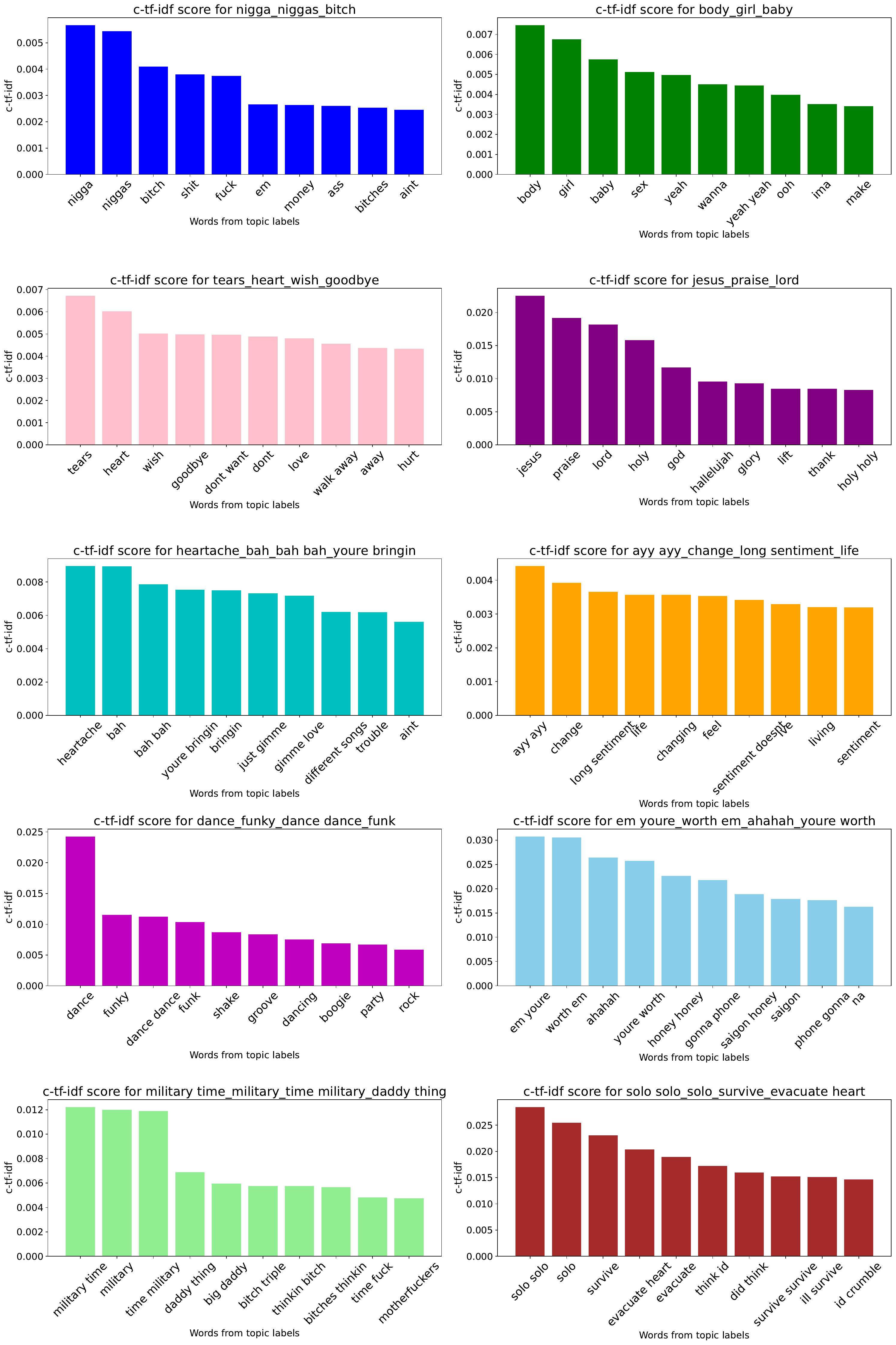}
    \caption{c-TF-IDF scores for words in the top 10 topics}
    \label{fig:c-tf-idf_appendix}
\end{figure*}

As shown in Figure \ref{fig:c-tf-idf_appendix}, the topic labels are derived by selecting words with the highest c-TF-IDF scores, which are identified by the BERTopic model~\cite{grootendorst2022bertopic}. Unlike traditional TF-IDF, c-TF-IDF computes word importance at the \textbf{cluster level} rather than the document level~\cite{juan2003, grootendorst2022bertopic}. This method ensures that the most representative and distinguishing terms for each topic are highlighted, facilitating the interpretation of thematic structures within the dataset.


\begin{table*}[]
    \centering
    \scalebox{0.9}{
    \begin{tabular}{p{0.35\linewidth} | p{0.65\linewidth}}
    \hline
    \textbf{Target Set} & \textbf{Words} \\
    \hline
        Pleasant & "friend", "joy", "wonderful", "vacation", "love", "honest", "honor", "pleasure", "loyal", "family", "peace", "heaven", "cheer", "freedom", "diploma", "gentle", "happy", "paradise", "diamond", "laughter", "sunrise", "gift", "health", "rainbow", "caress", "lucky", "miracle" \\
        Unpleasant & "terrible", "prison", "divorce", "war", "poverty", "sickness", "abuse", "tragedy", "hatred", "crash", "accident", "poison", "nasty", "awful", "grief", "disaster", "stink", "pollute", "ugly", "rotten", "filth", "failure", "bomb", "horrible", "jail", "kill", "cancer", "death", "murder", "evil", "vomit", "agony", "assault" \\
        Appearance words & "sensual", "thin", "handsome", "feeble", "bald", "fashionable", "slim", "gorgeous", "fat", "plump", "muscular", "pretty", "strong", "weak", "ugly", "slender", "homely", "healthy", "blushing", "athletic", "voluptuous", "stout", "beautiful", "alluring", "attractive" \\
        Intelligence words & "intelligent", "venerable", "adaptable", "reflective", "thoughtful", "resourceful", "genius", "logical", "smart", "astute", "judicious", "imaginative", "intuitive", "shrewd", "ingenious", "apt", "precocious", "inventive", "analytical", "inquiring", "inquisitive", "discerning", "brilliant", "clever", "wise" \\
        Strength words & "potent", "bold", "leader", "strong", "triumph", "command", "shout", "winner", "dominant", "power", "succeed", "confident", "dynamic", "loud", "assert" \\
        Weakness words & "wispy", "loser", "failure", "timid", "lose", "weak", "weakness", "shy", "surrender", "follow", "fragile", "withdraw", "vulnerable", "yield", "afraid" \\
        \hline
    \end{tabular}}
    \caption{List of target sets used for SC-WEAT analysis. These sets were chosen from the word sets curated by~\citet{betti2023large}, who compiled it from two different sources~\citep{Caliskan2017semantics,chaloner2019measuring}.} 
    \label{tab:target_set_appendix}
\end{table*}

\begin{table*}
\centering
    \scalebox{0.9}{
    \begin{tabular}{p{0.2\linewidth} | p{0.8\linewidth}}
    \hline
    \textbf{Attribute Set} & \textbf{Words} \\
    \hline
        Female & "aunt", "auntie", "daughter", "daughter-in-law", "female", "gal", "girl", "girlfriend", "grandmother", "grandmother-in-law", "her", "hers", "lady", "madam", "mama", "miss", "mom", "mother", "niece", "queen", "she", "sis", "sister", "wife", "woman" \\
        Male & "boy", "boyfriend", "brother", "dad", "father", "father-in-law", "grandfather", "grandpa", "guy", "he", "him", "his", "husband", "king", "male", "man", "nephew", "papa", "sir", "son", "son-in-law", "uncle"\\
    \hline
    \end{tabular}}
    \caption{List of attribute sets used for SC-WEAT analysis} 
    \label{tab:attribute_set_appendix} 
\end{table*}

\end{document}